\pdfoutput=1

\documentclass[11pt]{article}

\usepackage{naacl2021}

\usepackage{times}
\usepackage{latexsym}

\usepackage[T1]{fontenc}

\usepackage[utf8]{inputenc}

\usepackage{microtype}

\usepackage{url}
\usepackage{booktabs}       
\usepackage{amsfonts}       
\usepackage{nicefrac}       
\usepackage{caption}
\usepackage{subcaption}
\usepackage{microtype}      
\usepackage{graphicx}
\usepackage{amsmath, amsmath}
\usepackage{multirow}
\usepackage{lipsum}
\usepackage{breqn}
\usepackage{empheq}
\usepackage{multirow}
\usepackage{bbm}
\def\ie{{\em i.e.,~}}
\def\eg{{\em e.g.,~}}

\usepackage{xcolor}
\newcommand{\modify}[1]{{\color{black}{#1}}}

\usepackage{todonotes}
\usepackage{hyperref}
\definecolor{darktaupe}{rgb}{0.28, 0.24, 0.2}

\usepackage{adjustbox}
\usepackage{array}

\newcolumntype{R}[2]{%
    >{\adjustbox{angle=#1,lap=\width-(#2)}\bgroup}%
    l%
    <{\egroup}%
}

\title{Supporting Clustering with Contrastive Learning}

\author{Dejiao Zhang$^1$ \quad Feng Nan$^1$ \quad Xiaokai Wei$^1$ \quad Shang-Wen Li$^1$  \quad  Henghui Zhu$^1$ \\ \quad   \quad \textbf{Kathleen McKeown$^{1,2}$} \quad
\textbf{Ramesh Nallapati$^1$} \quad 
\textbf{Andrew O. Arnold$^1$}\quad \textbf{Bing Xiang$^1$} \\
  $^1$AWS AI \qquad  $^2$Columbia University\\
  \texttt{dejiaoz,nanfen,xiaokaiw,shangwel,henghui} \\
  \texttt{mckeownk,rnallapa,anarnld,bxiang@amazon.com}
  }

\begin{document}
\maketitle

\begin{abstract}
Unsupervised clustering aims at discovering the semantic categories of data according to some distance measured in the representation space. However, different categories often overlap with each other in the representation space at the beginning of the learning process, which poses a significant challenge for distance-based clustering in achieving good separation between different categories. To this end, we propose Supporting Clustering with Contrastive Learning (SCCL) -- a novel framework to leverage contrastive learning to promote better separation. We assess the performance of SCCL on short text clustering and show that SCCL significantly advances the state-of-the-art results on most 
benchmark datasets with $3\%-11\%$ improvement on Accuracy and $4\%-15\%$ improvement on Normalized Mutual Information. Furthermore, our quantitative analysis demonstrates the effectiveness of SCCL in leveraging the strengths of both bottom-up instance discrimination and  top-down clustering
to achieve better intra-cluster and inter-cluster distances when evaluated with the ground truth cluster labels\footnote{Our code is available at  \url{https://github.com/amazon-research/sccl}.}.
\end{abstract}

\section{Introduction}
\begin{figure}[htbp]
    \centering
    \includegraphics[scale=0.27]{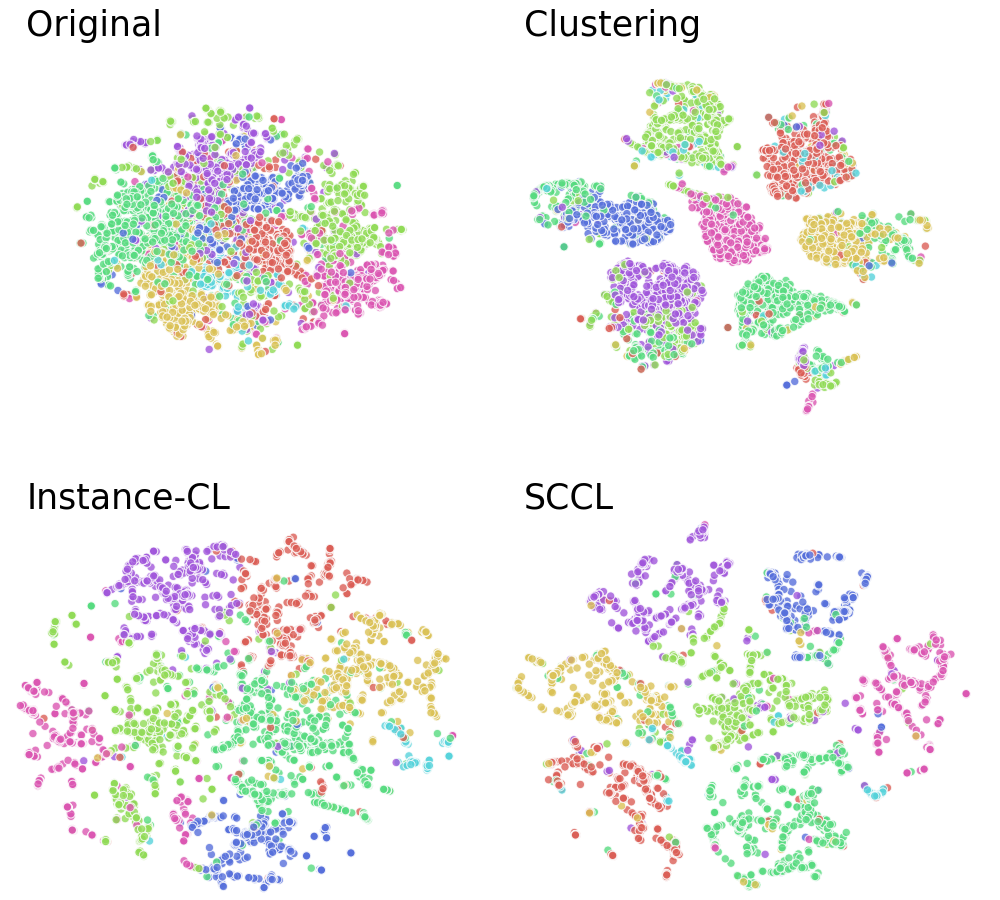}
    \caption{TSNE visualization of the embedding space learned on SearchSnippets using Sentence Transformer \citep{reimers-2019-sentence-bert} as backbone. Each color indicates a ground truth semantic category.
    }
    \label{fig:tsne_illustration}
\end{figure}
Clustering, one of the most fundamental challenges in unsupervised learning, has been widely studied for decades. 
Long established clustering methods
such as K-means \citep{macqueen1967some, lloyd1982least} and Gaussian Mixture Models \citep{celeux1995gaussian} rely on distance measured in the data space, which tends to be ineffective for high-dimensional data. 
On the other hand, deep neural networks are gaining momentum as an effective way to map data to a low dimensional and hopefully better separable representation space.

Many recent research efforts focus on integrating clustering with deep representation learning by optimizing a clustering objective defined in the representation space \citep{xie2016unsupervised,jiang2016variational,zhang2017deep,shaham2018spectralnet}. 
Despite promising improvements, the clustering performance is still inadequate, 
especially in the presence of complex data with a large number of clusters. As illustrated in Figure \ref{fig:tsne_illustration}, one possible reason is that, even with a deep neural network, data still has significant overlap across categories before clustering starts. Consequently, the clusters learned by optimizing various distance or similarity based clustering objectives suffer from poor purity.

On the other hand, Instance-wise Contrastive Learning (Instance-CL)  \citep{wu2018unsupervised,bachman2019learning, he2020momentum, chen2020simple, xchen2020improved} has recently achieved remarkable success in self-supervised learning. Instance-CL usually optimizes on an auxiliary set obtained by data augmentation. As the name suggests,  a contrastive loss is then adopted to pull together samples augmented from the same instance in the original dataset while pushing apart those from different ones. Essentially, Instance-CL disperses different instances apart while implicitly bringing similar instances together to some extent (see Figure \ref{fig:tsne_illustration}). This beneficial property can be leveraged to support clustering by scattering apart the overlapped categories.
Then clustering, thereby better separates different clusters while tightening each cluster by explicitly bringing samples in that cluster together.

To this end, we propose Supporting Clustering with Contrastive Learning (SCCL) by jointly optimizing a top-down clustering loss with a bottom-up instance-wise contrastive loss. 
We assess the performance of SCCL on short text clustering, which has become increasingly important due to the popularity of social media such as Twitter and Instagram. It benefits many real-world applications, including topic discovery \citep{kim2013discovering},  recommendation \citep{bouras2017improving}, and visualization \citep{sebrechts1999visualization}. However, the weak signal caused by noise and sparsity poses a significant challenge for clustering short texts. Although some improvement has been achieved by leveraging shallow neural networks to enrich the representations \citep{xu2017self, hadifar2019self}, there is still large room for improvement.

We address this challenge with our SCCL model. Our main contributions are the following:
\begin{itemize}
        \item We propose a novel end-to-end framework for unsupervised clustering, which advances the state-of-the-art results on various short text clustering datasets by a large margin. Furthermore, our model is much simpler than the existing deep neural network based short text clustering approaches that often require multi-stage independent training.  
    \item We provide in-depth analysis and demonstrate how SCCL effectively combines the top-down clustering with the bottom-up instance-wise contrastive learning to achieve better inter-cluster distance and intra-cluster distance.
    \item We explore various text augmentation techniques for SCCL, showing that, unlike the image domain \citep{chen2020simple}, using composition of augmentations is not always beneficial in the text domain. 
\end{itemize}

\begin{figure*}[htbp]
    \centering
    \includegraphics[scale=0.58]{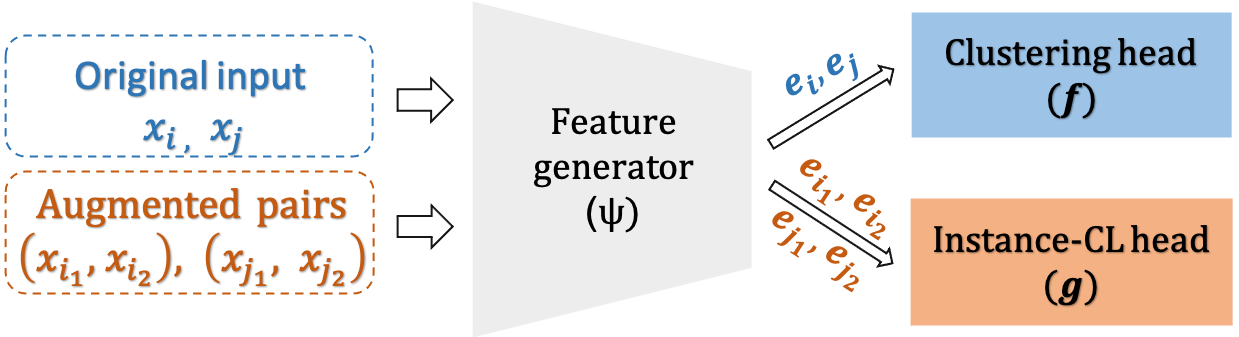}
    \caption{Training framework SCCL. During training, we jointly optimize a clustering loss over the original data instances and an instance-wise contrastive loss over the associated augmented pairs. }
    \label{fig:model}
\end{figure*}

\section{Related Work}
\paragraph{Self-supervised learning}{
Self-supervised learning has recently become prominent in providing effective representations for many downstream tasks. 
Early work focuses on solving different artificially designed pretext tasks, such as predicting masked tokens \citep{devlin2019bert},
generating future tokens \citep{radford2018improving}, or denoising corrupted tokens \citep{lewis2019bart} for textual data, and predicting colorization \citep{zhang2016colorful}, rotation \citep{gidaris2018unsupervised}, or relative patch position \citep{doersch2015unsupervised} for image data. Nevertheless, the resulting representations are tailored to the 
specific pretext tasks with limited generalization.

Many recent successes are largely driven by instance-wise contrastive learning. Inspired by the pioneering work of \citet{becker1992self,bromley1994signature}, Instance-CL treats each data instance and its augmentations as an independent class and tries to pull together the representations within each class while pushing apart different classes
\citep{dosovitskiy2014discriminative, oord2018representation, bachman2019learning,he2020momentum, chen2020simple, xchen2020improved}. Consequently,  different instances are well-separated in the learned embedding space with local invariance 
being preserved for each instance. 

Although Instance-CL may implicitly group similar instances together \citep{wu2018unsupervised}, it pushes representations apart as long as they are from different original instances, regardless of their semantic similarities. Thereby, the implicit grouping effect of Instance-CL is less stable and more data-dependent, giving rise to worse representations in some cases \citep{khosla2020supervised,li2020prototypical,purushwalkam2020demystifying}.
}   

\paragraph{Short Text Clustering}{Compared with the general text clustering problem, short text clustering comes with its own challenge due to the weak signal contained in each instance. In this scenario, BoW and TF-IDF often yield very sparse representation vectors that lack expressive ability. To remedy this issue, some early work leverages neural networks to enrich the representations \citep{xu2017self, hadifar2019self}, where  word embeddings \citep{mikolov2013distributed, arora2016simple} are adopted to further enhance the performance. 

However, the above approaches divide the learning process into multiple stages, each requiring independent optimization. On the other hand, despite the tremendous successes achieved by contextualized word embeddings \citep{peters2018deep, devlin2019bert, radford2018improving, reimers2019sentence}, 
they have been left largely unexplored for short text clustering. In this work, we leverage the pretrained transformer as the backbone, which is optimized in an end-to-end fashion. As demonstrated in Section \ref{sec:numerical}, we advance the state-of-the-art results on most benchmark datasets with $3\%-11\%$ improvement on Accuracy and $4\%-15\%$ improvement on NMI.
}

\section{Model}
\label{sec:model}
We aim at developing a joint model that leverages the beneficial properties of
Instance-CL to improve unsupervised clustering. As illustrated in Figure \ref{fig:model}, our model consists of three components. A neural network $\psi(\cdot)$ first maps the input data to the representation space, which is then followed by two different heads $g(\cdot)$ and $f(\cdot)$ where the contrastive loss and the clustering loss are applied, respectively. Please refer to Section \ref{sec:numerical} for details. 

Our data consists of both the original and the augmented data. Specifically, for a randomly sampled minibatch $\mathcal{B}=\{x_i\}_{i=1}^{M}$,  we randomly generate a pair of augmentations for each data instance in $\mathcal{B}$, yielding an augmented batch $\mathcal{B}^a$ with size $2M$, denoted as $\mathcal{B}^a =\{\tilde{x}_i\}_{i=1}^{2M}$. 

\subsection{Instance-wise Contrastive Learning}
\label{subsec:Instance-CL}
For each minibatch $\mathcal{B}$, the Instance-CL loss is defined on the augmented pairs in $\mathcal{B}^a$. Let $i^1\in\{1,\dots,2M\}$ denote the index of an arbitrary instance in augmented set $\mathcal{B}^a$, and let $i^2\in\{1,\dots,2M\}$ be the index of the other instance in $\mathcal{B}^a$ augmented from the same instance in the original set $\mathcal{B}$. We refer to $\tilde{x}_{i^1}, \tilde{x}_{i^2}\in\mathcal{B}^a$ as a \textit{positive} pair, while treating the other $2M$-2 examples in $\mathcal{B}^a$ as \textit{negative} instances regarding this positive pair. Let $\tilde{z}_{i^1}$ and $\tilde{z}_{i^{2}}$ be the corresponding outputs of the head $g$, \ie $\tilde{z}_{j}=g(\psi(\tilde{x}_j)), j=i^1, i^2$. Then for $\tilde{x}_{i^1}$, we try to separate $\tilde{x}_{i^2}$ apart from all negative instances in $\mathcal{B}^a$ by minimizing the following
\begin{align}
    \ell_{i^1}^{I} = -\log\frac{\exp(\text{sim}(\tilde{z}_{i^1}, \tilde{z}_{i^2})/\tau)}{\sum_{j=1}^{2M}\mathbbm{1}_{j\neq i^1} \cdot \exp(\text{sim}(\tilde{z}_{i^1}, \tilde{z}_j)/\tau)}\;. 
\end{align}
Here $\mathbbm{1}_{j\neq i^1}$ is an indicator function and $\tau$ denotes the temperature parameter which we set as $0.5$.
Following \citet{chen2020simple}, we choose $\text{sim}(\cdot)$ as the dot product between a pair of normalized outputs, \ie $\text{sim}(\tilde{z}_i, \tilde{z}_j) = \tilde{z}_i^T\tilde{z}_j / \|\tilde{z}_i\|_2 \|\tilde{z}_j\|_2$. 

The Instance-CL loss is then averaged over all instances in $\mathcal{B}^a$, 
\begin{align}
    \mathcal{L}_{\text{Instance-CL}} = \sum_{i=1}^{2M} \ell_{i}^{I} /2M \;.
    \label{eq:instance-CL}
\end{align}
To explore the above contrastive loss in the text domain, we explore three different augmentation strategies in Section \ref{subsec:var_aug}, where we find contextual augmenter  \citep{kobayashi2018contextual,ma2019nlpaug} consistently performs better than the other two.  

\subsection{Clustering}
\label{subsec:clustering}
We simultaneously encode the semantic categorical structure into the representations via unsupervised clustering. 
Unlike Instance-CL, clustering focuses on the high-level semantic concepts and tries to bring together instances from the same semantic category together. 
Suppose our data consists of $K$ semantic categories, and each category is characterized by its centroid in the representation space, denoted as $\mu_k, k\in\{1,\dots,K\}$. Let $e_j=\psi(x_j)$ denote the representation of instance $x_j$ in the original set $\mathcal{B}$. Following \citet{maaten2008visualizing}, we use the Student's t-distribution to compute the probability of assigning $x_j$ to the $k^{th}$ cluster, 
\begin{align}
    q_{jk} = \frac{\left(1 + \|e_j -\mu_k\|_2^2/\alpha \right)^{-\frac{\alpha+1}{2}}}{\sum_{k'=1}^{K} \left(1 + \|e_j -\mu_{k'}\|_2^2/\alpha\right)^{-\frac{\alpha+1}{2}}}\;.
    \label{eq:cluster_assignment}
\end{align}
Here $\alpha$ denotes the degree of freedom of the Student’s t-distribution. \modify{Without explicit mention, we follow \citet{maaten2008visualizing} by setting $\alpha=1$ in this paper}.

We use a linear layer, \ie the clustering head in Figure \ref{fig:model}, to approximate the centroids of each cluster, and we iteratively refine it by leveraging an auxiliary distribution proposed by \citet{xie2016unsupervised}. Specifically, let $p_{jk}$ denote the auxiliary probability defined as 
\begin{align}
    p_{jk} = \frac{q_{jk}^2/f_k}{\sum_{k'}q_{jk}^2/f_{k'}}\;.
    \label{eq:target}
\end{align}
Here $f_k = \sum_{j=1}^{M}q_{jk}, k=1, \dots,K$ can be interpreted as the soft cluster frequencies approximated within a minibatch. This target distribution first sharpens the soft-assignment probability $q_{jk}$ by raising it to the second power, and then normalizes it by the associated cluster frequency. By doing so, we encourage learning from high confidence cluster assignments and simultaneously combating the bias caused by imbalanced clusters.

We push the cluster assignment probability towards the target distribution by optimizing the KL divergence between them, 
\begin{align}
    \ell_j^C = \text{KL}\left[p_j \| q_j\right] = \sum_{k=1}^K p_{jk} \log \frac{p_{jk}}{q_{jk}} \;.
    \label{eq:KL_cluster_instance}
\end{align}
The clustering objective is then followed as 
\begin{align}
\mathcal{L}_{\text{Cluster}} = \sum_{j=1}^{M}  \ell_j^C / M
\label{eq:KL_cluster}
\end{align}
This clustering loss is first proposed in \citet{xie2016unsupervised} and later adopted by \citet{hadifar2019self} for short text clustering. However, they both require expensive layer-wise pretraining of the neural network, and update the target distribution (Eq \eqref{eq:target}) through carefully chosen intervals that often vary across datasets.  In contrast, we simplify the learning process to end-to-end training with the target distribution being updated per iteration. 

\paragraph{Overall objective} In summary, our overall objective is,
\begin{align}
\mathcal{L} &= \mathcal{L}_{\text{Instance-CL}} + \eta\mathcal{L}_{\text{Cluster}} \nonumber \\
&= \sum_{j=1}^{M}  \ell_j^C/M + \eta \sum_{i=1}^{2M}\ell_i^I / 2M    \;.
\label{eq:main_obj}
\end{align}
$\ell_j^C$ and $\ell_i^I$ are defined in Eq \eqref{eq:KL_cluster_instance} and Eq \eqref{eq:instance-CL}, respectively. 
\modify{$\eta$ balances between the contrastive loss and the clustering loss of SCCL, which we set as $10$ in  Section \ref{sec:numerical} for simplicity. Also noted that, the clustering loss is optimized over the original data only. Alternatively, we can also leverage the augmented data to enforce local consistency of the cluster assignments for each instance. We discuss this further in Appendix \ref{appendix_cluster}.}

\section{Numerical Results}
\label{sec:numerical}
\paragraph{Implementation}
We implement our model in PyTorch \citep{paszke2017automatic} with the Sentence Transformer library  \citep{reimers-2019-sentence-bert}.  We choose \textit{distilbert-base-nli-stsb-mean-tokens} as the backbone, followed by a linear clustering head $(f)$ of size $768\times K$ with $K$ indicating the number of clusters. For the contrastive loss, we optimize an MLP $(g)$ with one hidden layer of size 768, and output vectors of size 128. Figure \ref{fig:model} provides an illustration of our model. The detailed experimental setup is provided in Appendix \ref{appendix:exp_setup}. We, as in the previous work \citet{xu2017self, hadifar2019self,rakib2020enhancement}, adopt Accuracy (ACC)  and Normalized Mutual Information (NMI) to evaluate different approaches.  

\begin{table*}[htbp]
  \begin{center}
    \begin{tabular}{lcccccccc}
      &\multicolumn{2}{c}{\textbf{AgNews}} 
      &\multicolumn{2}{c}{\textbf{SearchSnippets}} & \multicolumn{2}{c}{\textbf{StackOverflow}} & \multicolumn{2}{c}{\textbf{Biomedical}}  \\
      & ACC & NMI & ACC & NMI & ACC & NMI & ACC & NMI \\
      \cline{2-9}
      BoW &27.6 &2.6 & 24.3&9.3 &18.5&14.0 &14.3 &9.2 \\
      TF-IDF &34.5 &11.9 &31.5 &19.2 &58.4 &58.7 &28.3 &23.2 \\
      {STCC} &- &- & 77.0 &63.2 &51.1 &49.0 &43.6 &38.1  \\
       Self-Train &- &- &77.1& 56.7& 59.8 &54.8 &\textbf{54.8}&\textbf{47.1} \\
      HAC-SD &81.8 &54.6 &82.7 &63.8 &64.8 &59.5 &40.1 &33.5 \\
      \hline
      \textcolor{black}{\textbf{SCCL}}&\textbf{88.2} &\textbf{68.2} &
      \textbf{85.2} &\textbf{71.1} &\textbf{75.5}&\textbf{74.5} &46.2 &41.5 \\ \\
     
      & \multicolumn{2}{c}{\textbf{GoogleNews-TS}} & \multicolumn{2}{c}{\textbf{GoogleNews-T}} & \multicolumn{2}{c}{\textbf{GoogleNews-S}}&
      \multicolumn{2}{c}{\textbf{Tweet}}  \\
      & ACC & NMI & ACC & NMI & ACC & NMI & ACC & NMI \\
      \cline{2-9}
      BoW &57.5 &81.9 &49.8 &73.2 &49.0 &73.5 &49.7 &73.6 \\
      TF-IDF &68.0 &88.9  &58.9 &79.3 &61.9 &83.0 &57.0 &80.7 \\
      STCC & -&- &-&- &-&- &-&- \\
      Self-Train & -&- &- &- &-&- &-&- \\
      HAC-SD &85.8 &88.0 &\textbf{81.8} &84.2 &80.6 &83.5&\textbf{89.6} &85.2\\
      \hline  
      \textcolor{black}{\textbf{SCCL}}& \textbf{89.8}&\textbf{94.9} &75.8 &\textbf{88.3}  & \textbf{83.1} &\textbf{90.4} &78.2 &\textbf{89.2}\\
    \end{tabular}
    \caption{Clustering results on eight short text datasets. Our results are averaged over five random runs. }
    \label{tab:SOTA_compare}
  \end{center}
\end{table*}

\paragraph{Datasets}
We assess the performance of the proposed SCCL model on eight benchmark datasets for short text clustering. Table \ref{tab:datastats} provides an overview of the main statistics, and the details of each dataset are as follows. 

\begin{itemize}
    \item \textbf{SearchSnippets} is extracted from web search snippets, which contains 12,340 snippets associated with 8 groups \citet{phan2008learning}.
    \item \textbf{StackOverflow} is a subset of the challenge data published by Kaggle\footnote{https://www.kaggle.com/c/predict-closed-questions-on-stackoverflow/download/train.zip}, where 20,000 question titles associated with 20 different categories are selected by \citet{xu2017self}.

    \item \textbf{Biomedical} is a subset of the PubMed data distributed by BioASQ\footnote{http://participants-area.bioasq.org}, where 20,000 paper titles from 20 groups are randomly selected by \citet{xu2017self}.

    \item \textbf{AgNews} is a subset of news
titles \citep{zhang2015text}, which contains 4 topics selected by \citet{rakib2020enhancement}. 
    \item \textbf{Tweet} consists of 2,472 tweets with 89 categories \citep{yin2016model}.

    \item \textbf{GoogleNews} contains titles and snippets of 11,109 news articles related to 152 events \citep{yin2016model}. Following \citep{rakib2020enhancement}, we name the full dataset as \textit{GoogleNews-TS}, and  \textit{GoogleNews-T} and \textit{GoogleNews-S} are obtained by extracting the titles and the snippets,  respectively. 
\end{itemize}
For each dataset, we use \textit{Contextual Augmenter} \citep{kobayashi2018contextual,ma2019nlpaug} to obtain the augmentation set, as it consistently outperforms the other options  explored in Section \ref{subsec:var_aug}.  

\begin{table}[!htbp]
  \begin{center}
    {\footnotesize
    \begin{tabular}{l|ccccc}
    \hline
    Dataset & $|V|$ &\multicolumn{2}{c}{\text{Documents}}  &\multicolumn{2}{c}{\text{Clusters}}\\
    & & $N^D$ & Len & $N^C$ & L/S \\
    \hline
    AgNews &21K &8000 &23 &4 & 1   \\
    StackOverflow& 15K &20000 &8 & 20 &1   \\
    Biomedical& 19K &20000 &13 &  20 &1  \\
    SearchSnippets& 31K &12340 &18 & 8 & 7   \\
    GooglenewsTS& 20K &11109 & 28 & 152 & 143   \\
    GooglenewsS&18K &11109 & 22 & 152 & 143  \\
    GooglenewsT& 8K &11109 & 6 & 152 & 143  \\
    Tweet& 5K &2472 & 8 & 89 & 249 \\
      \hline
    \end{tabular}
    \caption{Dataset statistics. $|V|$: the vocabulary size; $N^D$: number of short text documents; Len: average number of words in each document; $N^C$ number of clusters; L/S: the ratio of the size of the largest cluster to that of the smallest cluster. }
    \label{tab:datastats}}
  \end{center}
\end{table}

\begin{figure*}[htbp]
    \centering
    \includegraphics[scale=0.38]{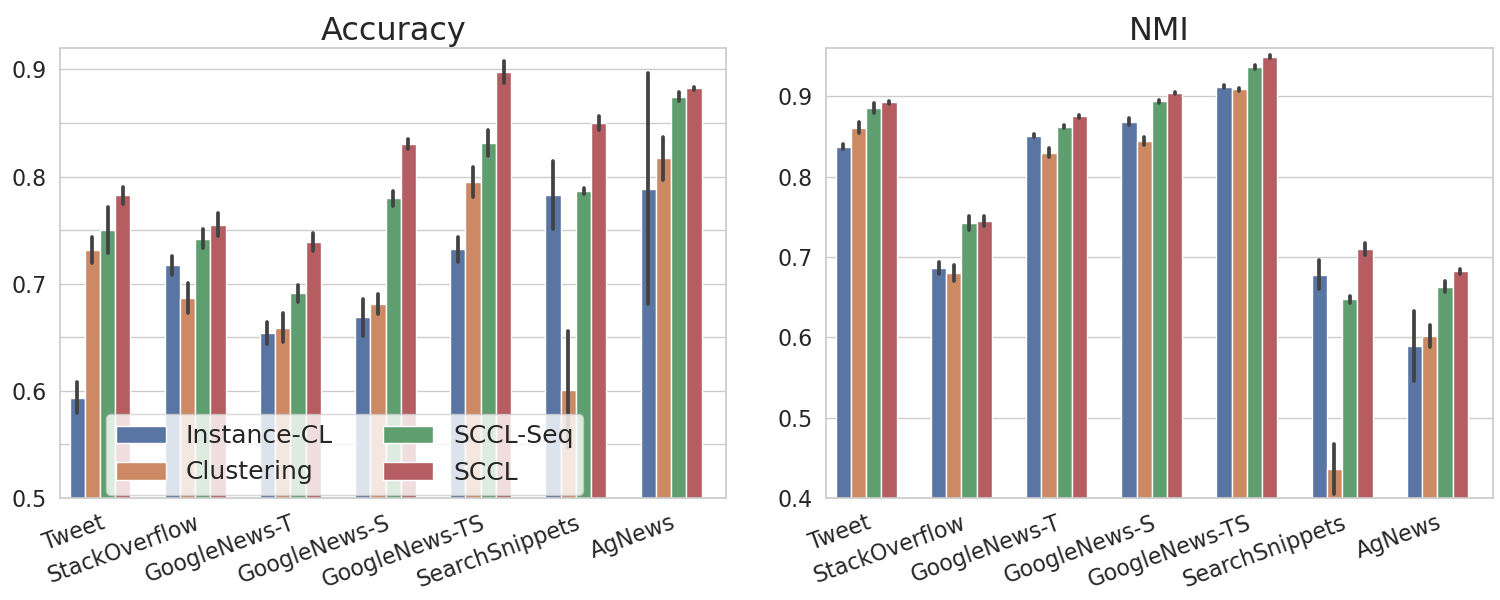}
    \caption{Ablation study of SCCL.  In SCCL-Seq, we first train the model using Instance-CL, and then optimize the clustering objective. We exclude Biomedical for better visualization, full plot can be found in Appendix \ref{appdix:ablation}.}
    \label{fig:ablation_study}
\end{figure*}

\subsection{Comparison with State-of-the-art}
We first demonstrate that our model can achieve state-of-the-art or highly competitive performance on short text clustering. For comparison, we consider the following baselines. 
\begin{itemize}
    \item \textbf{STCC} \citep{xu2017self} consists of three independent stages. For each dataset, it first pretrains a word embedding on a large in-domain corpus using the Word2Vec method \citep{mikolov2013efficient}. A convolutional neural network is then optimized to further enrich the representations that are fed into K-means for the final stage clustering. 
    \item\textbf{Self-Train} \citep{hadifar2019self} enhances the pretrained word embeddings in \citet{xu2017self} using SIF \citep{arora2016simple}. Following \citet{xie2016unsupervised}, it adopts an auto-encoder obtained by layer-wise pretraining  \citep{van2009learning}, which is then further tuned with a clustering objective same as that in Section \ref{subsec:clustering}. Both \citet{xie2016unsupervised} and \citet{hadifar2019self} update the target distribution through carefully chosen intervals that vary across datasets, while we update it per iteration yet still achieve significant improvement. 
    \item\textbf{HAC-SD} \citep{rakib2020enhancement}\footnote{They further boost the performance via an iterative classification trained with high-confidence pseudo labels extracted after each round of clustering. Since the iterative classification strategy is orthogonal to the clustering algorithms, we only evaluate against with their proposed clustering algorithm for fair comparison.} applies hierarchical agglomerative clustering on top of a sparse pairwise similarity matrix obtained by zeroing-out similarity scores lower than a chosen threshold value. 
  \item \textbf{BoW} \& \textbf{TF-IDF} are evaluated by applying K-means on top of the associated features with dimension being 1500. 
\end{itemize}

To demonstrate that our model is robust against the noisy input that often poses a significant challenge for short text clustering, we do not apply any pre-processing procedures on any of the eight datasets. 
In contrast, 
\modify{all baselines except BoW and TF-IDF}
considered in this paper either pre-processed the Biomedical dataset \citep{xu2017self, hadifar2019self} or all eight datasets by removing the stop words, punctuation, and converting the text to lower case \citep{rakib2020enhancement}.


We report the comparison results in Table \ref{tab:SOTA_compare}. Our SCCL model outperforms all baselines by a large margin on most datasets. Although we are lagging behind \citet{hadifar2019self} on Biomedical, SCCL still shows great promise considering the fact that Biomedical is much less related to the general domains on which the transformers are pretrained. In contrast, \citet{hadifar2019self} learn the word embeddings on a large in-domain biomedical corpus, followed by a layer-wise pretrained autoencoder to further enrich the representations. 

\modify{\citet{rakib2020enhancement} also shows better Accuracy on Tweet and GoogleNews-T, for which we hypothesize two reasons. First, both GoogleNews and Tweet have fewer training examples with much more clusters. Thereby, it's challenging for instance-wise contrast learning to manifest its advantages, which often requires a large training dataset. Second, as implied by the clustering perfermance evaluated on BoW and TF-IDF, clustering GoogleNews and Tweet is less challenging than clustering the other four datasets. Hence, by applying agglomerative clustering on the carefully selected pairwise similarities of the preprocessed data, \citet{rakib2020enhancement} can achieve good performance, especially when the text instances are very short, \ie Tweet and GoogleNews-T.  We also highlight the scalability of our model to large scale data, whereas agglomerative clustering often suffers from high computation complexity. We discuss this further in Appendix \ref{appendix_rakib}.}

\subsection{Ablation Study}
To better validate our model, we run ablations in this section.
For illustration, we name the clustering component described in Section \ref{subsec:clustering} as Clustering. Besides Instance-CL and Clustering,  we also evaluate SCCL against its sequential version (SCCL-Seq) where we first train the model with Instance-CL, and then optimize it with Clustering. 

As shown in Figure \ref{fig:ablation_study}, Instance-CL also groups semantically similar instances together. \modify{However, this grouping effect is implicit and data-dependent.}
In contrast, SCCL consistently outperforms both Instance-CL and Clustering by a large margin. 
Furthermore, SCCL also achieves better performance than its sequential version, SCCL-Seq. The result validates the effectiveness and importance of the proposed joint optimization framework in leveraging the strengths of both Instance-CL and Clustering to compliment each other.  

\begin{table*}[!htbp]
  \begin{center}
    \begin{tabular}{lccccccc}
    \textbf{Dataset}  &\multicolumn{3}{c}{\textbf{Accuracy}}& &\multicolumn{3}{c}{\textbf{NMI}} \\
    \cline{2-4} \cline{6-8} 
    &\textbf{WNet}  &\textbf{Para}
    &\textbf{Ctxt}& &\textbf{WNet}  &\textbf{Para}
    &\textbf{Ctxt}\\
    
    AgNews& 86.6 &        86.5 &        \textbf{88.2} &       &      66.0 &      65.2 &      \textbf{68.2} \\
    SearchSnippets& 78.1 &        83.7 &        \textbf{85.0} &       &      61.9 &      68.1 &      \textbf{71.0} \\
    StackOverflow& 69.1 &        73.3 &        \textbf{75.5} &       &      69.9 &      72.7 &      \textbf{74.5} \\
    Biomedical& 42.8 &        43.0 &        \textbf{46.2} &       &      38.0 &      39.5 &      \textbf{41.5} \\
    GooglenewsTS& 82.1 &        83.5 &        \textbf{89.8} &       &      92.1 &      92.9 &      \textbf{94.9} \\
    GooglenewsS& 73.0 &        75.3 &        \textbf{83.1} &       &      86.4 &      87.4 &      \textbf{90.4} \\
    GooglenewsT& 66.3 &        67.5 &        \textbf{73.9} &       &      83.4 &      83.6 &      \textbf{87.5} \\
    Tweet& 70.6 &        73.7 &        \textbf{78.2} &       &      86.2 &      86.4 &      \textbf{89.2} \\
    \end{tabular}
    \caption{Results of SCCL evaluated with different augmentation techniques: WordNet augmenter (\textbf{WNet}), paraphrase via back translation (\textbf{Para}), and contextual augmenter (\textbf{Ctxt}).
     Each technique is detailed in Section \ref{subsec:var_aug}.}
     \label{tab:varaug}
  \end{center}
\end{table*}

\subsubsection{SCCL leads to better separated and less dispersed clusters} 
\begin{figure}[htbp]
    \centering
        \includegraphics[scale=0.25]{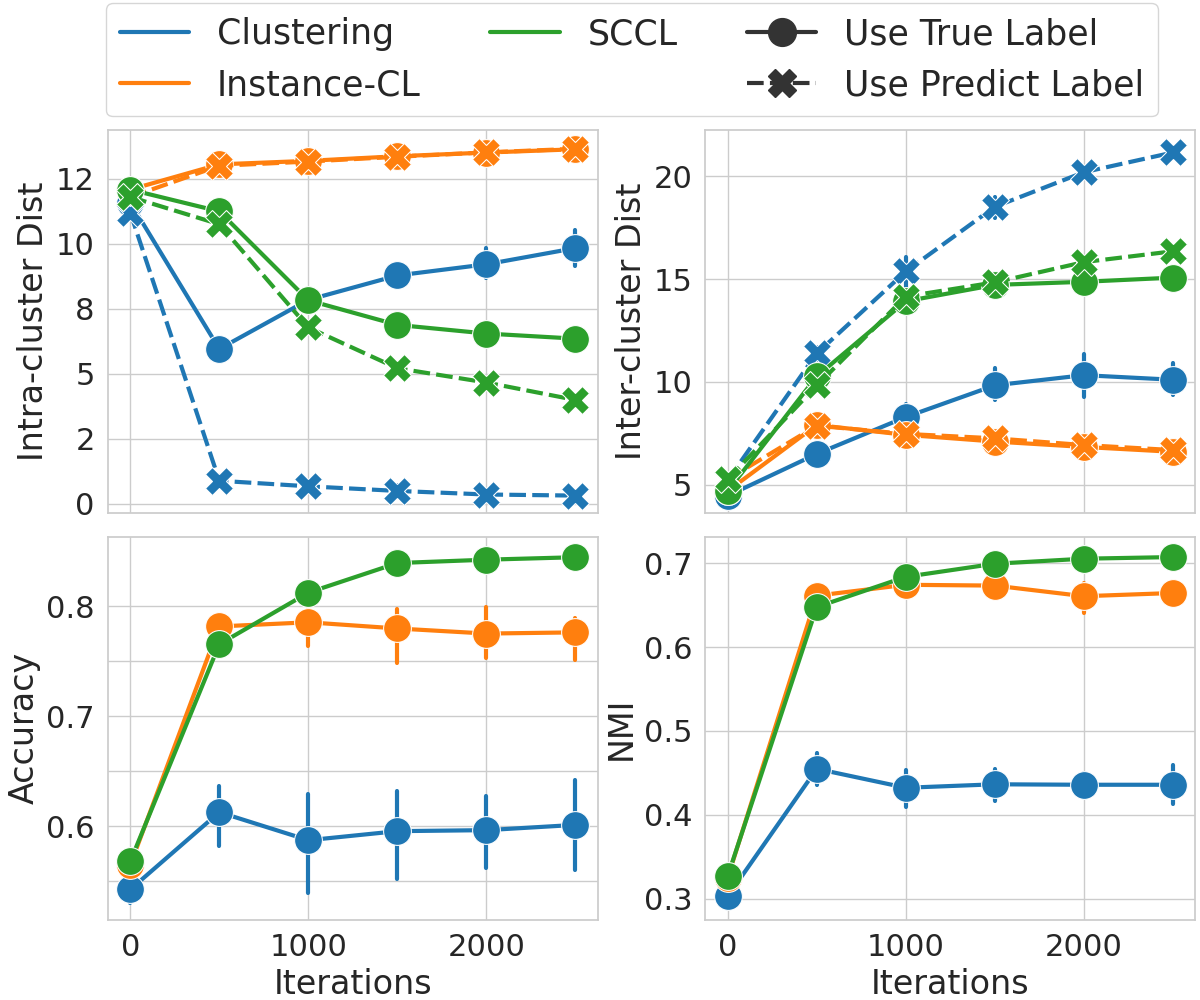}
    \caption{Cluster-level evaluation on SearchSnippets. Each plot is summarized over five random runs.}
        \label{fig:learn_stats}
\end{figure}

To further investigate what enables the better performance of SCCL,  we track both the intra-cluster distance and the inter-cluster distance evaluated in the representation space throughout the learning process. For a given cluster, the intra-cluster distance is the average distance between the centroid and all samples grouped into that cluster, and the inter-cluster distance is the distance to its closest neighbor cluster. In Figure \ref{fig:learn_stats}, we report each type of distance with its mean value obtained by averaging over all clusters, where the clusters are defined either regarding the ground truth labels (solid lines) or the labels predicted by the model (dashed lines). 

Figure \ref{fig:learn_stats} shows  
Clustering achieves smaller intra-cluster distance and larger inter-cluster distance when evaluated on the predicted clusters. It demonstrates the ability of Clustering to tight each self-learned cluster and separate different clusters apart. However, we observe the opposite when evaluated on the ground truth clusters,  along with poor Accuracy and NMI scores. One possible explanation is, data from different ground-truth clusters often have significant overlap in the embedding space before clustering starts (see upper left plot in Figure \ref{fig:tsne_illustration}), which makes it hard for our distance-based clustering approach to separate them apart effectively.

Although the implicit grouping effect allows Instance-CL attains better Accuracy and NMI scores, the resulting clusters are less apart from each other and each cluster is more dispersed, \modify{as indicated by the smaller inter-cluster distance and larger intra-cluster distance}. This result is unsurprising since Instance-CL only focuses on instance discrimination, which often leads to a more dispersed embedding space. 
In contrast, we leverage the strengths of both Clustering and Instance-CL to compliment each other. Consequently, 
Figure \ref{fig:learn_stats} shows SCCL leads to better separated clusters with each cluster being less dispersed.

\begin{figure*}[htbp]
    \centering
    \begin{subfigure}{0.49\textwidth}
        \centering
        \includegraphics[scale=0.4]{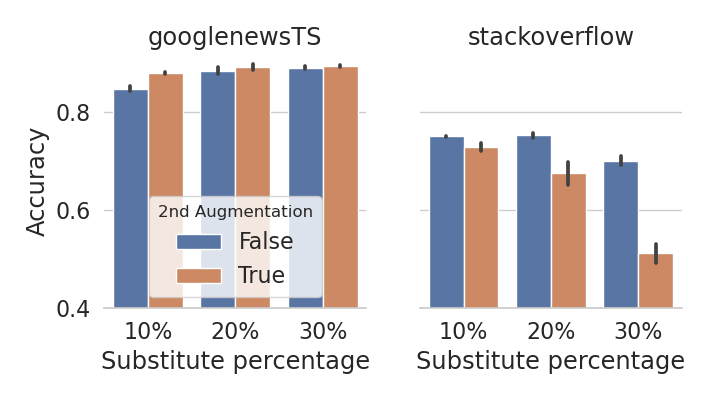}
    \end{subfigure}
    \begin{subfigure}{0.49\textwidth}
        \centering
         \includegraphics[scale=0.4]{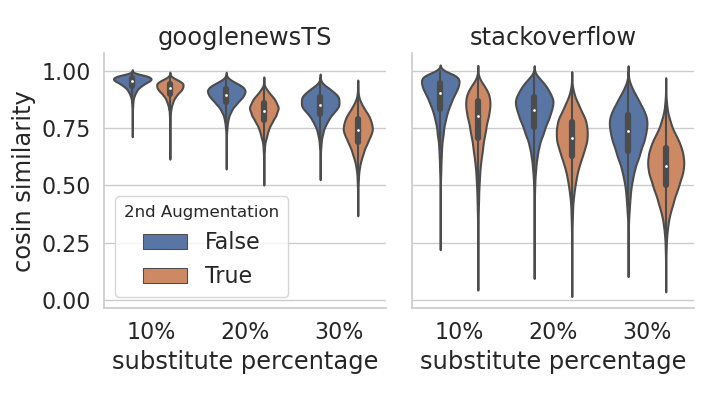}
    \end{subfigure}
    \caption{Impact of using composition of data augmentations. Either only \text{Contextual Augmenter} (\textcolor{blue}{blue}) is used, or \text{Contextual Augmenter} and \text{CharSwap Augmenter} are applied  sequentially (\textcolor{orange}{orange}).  \textbf{(Left)} Clustering accuracy versus variant augmentation strengths, the x-axis indicates the percentage of words in each instance being changed by the associated data augmentation technique. \textbf{(Right)} Distribution of the cosine similarity between the representations of each original text and its augmented pair at the beginning of training.}
    \label{fig:composition_of_augmentation}
\end{figure*}

\subsection{Data Augmentation}
\label{subsec:data_aug}
\subsubsection{Exploration of Data Augmentations}
\label{subsec:var_aug}
To study the impact of data augmentation, we explore three different unsupervised text augmentations: (1) \textit{WordNet Augmenter}\footnote{\url{https://github.com/QData/TextAttack}} transforms an input text by replacing its words with WordNet synonyms
\citep{morris2020textattack,ren2019generating}. (2) \textit{Contextual Augmenter}\footnote{\url{https://github.com/makcedward/nlpaug}} leverages the pretrained transformers to find top-n suitable words of the input text for insertion or substitution  \citep{kobayashi2018contextual,ma2019nlpaug}. We augment the data via word substitution, and we choose Bertbase and Roberta to generate the augmented pairs.
(3) \textit{Paraphrase via back translation}\footnote{ \url{https://github.com/pytorch/fairseq/tree/master/examples/paraphraser}} generates paraphrases of the input text by first translating it to another language (French) and then back to English. When translating back to English, we used the mixture of experts model \cite{shen2019mixture} to generate ten candidate paraphrases per input to increase diversity. 

For both \textit{WordNet Augmenter} and \textit{Contextual Augmenter}, we try three different settings by choosing the word substitution ratio of each text instance to $10\%$, $20\%$, and $30\%$, respectively. As for \textit{Paraphrase via back translation}, we compute the BLEU score between each text instance and its ten candidate paraphrases. We then select three pairs, achieving the highest, medium, and lowest BLEU scores, from the ten condidates of each instance. The best results\footnote{Please refer to Appendix \ref{appendix:var_aug} for details.} of each augmentation technique are summarized in Table \ref{tab:varaug}, where \textit{Contexual Augmenter} substantially outperforms the other two. We conjecture that this is due to both \textit{Contextual Augmenter} and SCCL leverage the pretrained transformers as backbones, which allows \textit{Contextual Augmenter} to generate more informative augmentations.

\subsubsection{Composition of Data Augmentations}
\label{subsec:comp_dataaug}
Figure \ref{fig:composition_of_augmentation} shows the impact of using composition of data augmentations, in which we explored \textit{Contextual Augmenter} and \textit{CharSwap Augmenter}\footnote{A simple technique that augments text by substituting, deleting, inserting, and swapping adjacent characters} \citep{morris2020textattack}.  As we can see, using composition of data augmentations does boost the performance of SCCL on GoogleNews-TS where the average number of words in each text instance is $28$ (see Table \ref{tab:datastats}).
However, we observe the opposite on StackOverflow where the average number of words in each instance is 8. This result differs from what has been observed in the image domain where using composition of data augmentations is crucial for contrastive learning to attain good performance. Possible explanations is that generating high-quality augmentations for textual data is more challenging, since changing a single word can invert the semantic meaning of the whole instance. This challenge is compounded when a second round of augmentation is applied on very short text instances, \eg StackOverflow. We further demonstrate this in Figure \ref{fig:composition_of_augmentation} (right), where the augmented pairs of StackOverflow largely diverge from the original texts in the representation space after the second round of augmentation.  

\section{Conclusion}
We have proposed a novel framework leveraging instance-wise contrastive learning to support unsupervised clustering. We thoroughly evaluate our model on eight benchmark short text clustering datasets, and show that our model either  substantially outperforms or performs highly comparably to the state-of-the-art methods. Moreover, we conduct ablation studies to better validate the effectiveness of our model. We demonstrate that, by integrating the strengths of both bottom-up instance discrimination and  top-down clustering, our model is capable of generating high-quality clusters with better intra-cluster and inter-clusters distances. Although we only evaluate our model on short text data, the proposed framework is generic and is expected to be effective for various kinds of text clustering problems. 


\modify{In this work, we explored different data augmentation strategies with extensive comparisons. However, due to the discrete nature of natural language, designing effective transformations for textual data is more challenging compared to the counterparts in the computer vision domain. One promising direction is leveraging the data mixing strategies \citep{zhang2017mixup} to either obtain stronger augmentations \citep{kalantidis2020hard} or alleviate the heavy burden on data augmentation  \citep{lee2020mix}. We leave this as future work.}

\bibliography{anthology}

\begin{thebibliography}{55}
\expandafter\ifx\csname natexlab\endcsname\relax\def\natexlab#1{#1}\fi

\bibitem[{Arora et~al.(2017)Arora, Liang, and Ma}]{arora2016simple}
Sanjeev Arora, Yingyu Liang, and Tengyu Ma. 2017.
\newblock A simple but tough-to-beat baseline for sentence embeddings.
\newblock In \emph{International Conference on Learning Representations}.

\bibitem[{Bachman et~al.(2019)Bachman, Hjelm, and
  Buchwalter}]{bachman2019learning}
Philip Bachman, R~Devon Hjelm, and William Buchwalter. 2019.
\newblock Learning representations by maximizing mutual information across
  views.
\newblock In \emph{Advances in Neural Information Processing Systems}, pages
  15535--15545.

\bibitem[{Becker and Hinton(1992)}]{becker1992self}
Suzanna Becker and Geoffrey~E Hinton. 1992.
\newblock Self-organizing neural network that discovers surfaces in random-dot
  stereograms.
\newblock \emph{Nature}, 355(6356):161--163.

\bibitem[{Bouras and Tsogkas(2017)}]{bouras2017improving}
Christos Bouras and Vassilis Tsogkas. 2017.
\newblock Improving news articles recommendations via user clustering.
\newblock \emph{International Journal of Machine Learning and Cybernetics},
  8(1):223--237.

\bibitem[{Bromley et~al.(1994)Bromley, Guyon, LeCun, S{\"a}ckinger, and
  Shah}]{bromley1994signature}
Jane Bromley, Isabelle Guyon, Yann LeCun, Eduard S{\"a}ckinger, and Roopak
  Shah. 1994.
\newblock Signature verification using a" siamese" time delay neural network.
\newblock In \emph{Advances in neural information processing systems}, pages
  737--744.

\bibitem[{Celeux and Govaert(1995)}]{celeux1995gaussian}
Gilles Celeux and G{\'e}rard Govaert. 1995.
\newblock Gaussian parsimonious clustering models.
\newblock \emph{Pattern recognition}, 28(5):781--793.

\bibitem[{Chen et~al.(2020{\natexlab{a}})Chen, Kornblith, Norouzi, and
  Hinton}]{chen2020simple}
Ting Chen, Simon Kornblith, Mohammad Norouzi, and Geoffrey Hinton.
  2020{\natexlab{a}}.
\newblock A simple framework for contrastive learning of visual
  representations.
\newblock \emph{arXiv preprint arXiv:2002.05709}.

\bibitem[{Chen et~al.(2020{\natexlab{b}})Chen, Fan, Girshick, and
  He}]{xchen2020improved}
Xinlei Chen, Haoqi Fan, Ross Girshick, and Kaiming He. 2020{\natexlab{b}}.
\newblock Improved baselines with momentum contrastive learning.
\newblock \emph{arXiv preprint arXiv:2003.04297}.

\bibitem[{Devlin et~al.(2019)Devlin, Chang, Lee, and
  Toutanova}]{devlin2019bert}
Jacob Devlin, Ming-Wei Chang, Kenton Lee, and Kristina Toutanova. 2019.
\newblock Bert: Pre-training of deep bidirectional transformers for language
  understanding.
\newblock In \emph{Proceedings of the 2019 Conference of the North American
  Chapter of the Association for Computational Linguistics: Human Language
  Technologies, Volume 1 (Long and Short Papers)}, pages 4171--4186.

\bibitem[{Doersch et~al.(2015)Doersch, Gupta, and
  Efros}]{doersch2015unsupervised}
Carl Doersch, Abhinav Gupta, and Alexei~A Efros. 2015.
\newblock Unsupervised visual representation learning by context prediction.
\newblock In \emph{Proceedings of the IEEE international conference on computer
  vision}, pages 1422--1430.

\bibitem[{Dosovitskiy et~al.(2014)Dosovitskiy, Springenberg, Riedmiller, and
  Brox}]{dosovitskiy2014discriminative}
Alexey Dosovitskiy, Jost~Tobias Springenberg, Martin Riedmiller, and Thomas
  Brox. 2014.
\newblock Discriminative unsupervised feature learning with convolutional
  neural networks.
\newblock In \emph{Advances in neural information processing systems}, pages
  766--774.

\bibitem[{Gidaris et~al.(2018)Gidaris, Singh, and
  Komodakis}]{gidaris2018unsupervised}
Spyros Gidaris, Praveer Singh, and Nikos Komodakis. 2018.
\newblock Unsupervised representation learning by predicting image rotations.
\newblock \emph{arXiv preprint arXiv:1803.07728}.

\bibitem[{Giorgi et~al.(2020)Giorgi, Nitski, Bader, and
  Wang}]{giorgi2020declutr}
John~M Giorgi, Osvald Nitski, Gary~D Bader, and Bo~Wang. 2020.
\newblock Declutr: Deep contrastive learning for unsupervised textual
  representations.
\newblock \emph{arXiv preprint arXiv:2006.03659}.

\bibitem[{Hadifar et~al.(2019)Hadifar, Sterckx, Demeester, and
  Develder}]{hadifar2019self}
Amir Hadifar, Lucas Sterckx, Thomas Demeester, and Chris Develder. 2019.
\newblock A self-training approach for short text clustering.
\newblock In \emph{Proceedings of the 4th Workshop on Representation Learning
  for NLP (RepL4NLP-2019)}, pages 194--199.

\bibitem[{He et~al.(2020)He, Fan, Wu, Xie, and Girshick}]{he2020momentum}
Kaiming He, Haoqi Fan, Yuxin Wu, Saining Xie, and Ross Girshick. 2020.
\newblock Momentum contrast for unsupervised visual representation learning.
\newblock In \emph{Proceedings of the IEEE/CVF Conference on Computer Vision
  and Pattern Recognition}, pages 9729--9738.

\bibitem[{Jiang et~al.(2016)Jiang, Zheng, Tan, Tang, and
  Zhou}]{jiang2016variational}
Zhuxi Jiang, Yin Zheng, Huachun Tan, Bangsheng Tang, and Hanning Zhou. 2016.
\newblock Variational deep embedding: An unsupervised and generative approach
  to clustering.
\newblock \emph{arXiv preprint arXiv:1611.05148}.

\bibitem[{Kalantidis et~al.(2020)Kalantidis, Sariyildiz, Pion, Weinzaepfel, and
  Larlus}]{kalantidis2020hard}
Yannis Kalantidis, Mert~Bulent Sariyildiz, Noe Pion, Philippe Weinzaepfel, and
  Diane Larlus. 2020.
\newblock Hard negative mixing for contrastive learning.
\newblock \emph{arXiv preprint arXiv:2010.01028}.

\bibitem[{Khosla et~al.(2020)Khosla, Teterwak, Wang, Sarna, Tian, Isola,
  Maschinot, Liu, and Krishnan}]{khosla2020supervised}
Prannay Khosla, Piotr Teterwak, Chen Wang, Aaron Sarna, Yonglong Tian, Phillip
  Isola, Aaron Maschinot, Ce~Liu, and Dilip Krishnan. 2020.
\newblock Supervised contrastive learning.
\newblock \emph{arXiv preprint arXiv:2004.11362}.

\bibitem[{Kim et~al.(2013)Kim, Lee, and Kyeong}]{kim2013discovering}
Hwi-Gang Kim, Seongjoo Lee, and Sunghyon Kyeong. 2013.
\newblock Discovering hot topics using twitter streaming data social topic
  detection and geographic clustering.
\newblock In \emph{2013 IEEE/ACM International Conference on Advances in Social
  Networks Analysis and Mining (ASONAM 2013)}, pages 1215--1220. IEEE.

\bibitem[{Kingma and Ba(2015)}]{KingmaB14}
Diederik~P. Kingma and Jimmy Ba. 2015.
\newblock \href {http://arxiv.org/abs/1412.6980} {Adam: {A} method for
  stochastic optimization}.
\newblock In \emph{3rd International Conference on Learning Representations,
  {ICLR} 2015, San Diego, CA, USA, May 7-9, 2015, Conference Track
  Proceedings}.

\bibitem[{Kobayashi(2018)}]{kobayashi2018contextual}
Sosuke Kobayashi. 2018.
\newblock Contextual augmentation: Data augmentation by words with paradigmatic
  relations.
\newblock \emph{arXiv preprint arXiv:1805.06201}.

\bibitem[{Lee et~al.(2020)Lee, Zhu, Sohn, Li, Shin, and Lee}]{lee2020mix}
Kibok Lee, Yian Zhu, Kihyuk Sohn, Chun-Liang Li, Jinwoo Shin, and Honglak Lee.
  2020.
\newblock i-mix: A strategy for regularizing contrastive representation
  learning.
\newblock \emph{arXiv preprint arXiv:2010.08887}.

\bibitem[{Lewis et~al.(2019)Lewis, Liu, Goyal, Ghazvininejad, Mohamed, Levy,
  Stoyanov, and Zettlemoyer}]{lewis2019bart}
Mike Lewis, Yinhan Liu, Naman Goyal, Marjan Ghazvininejad, Abdelrahman Mohamed,
  Omer Levy, Ves Stoyanov, and Luke Zettlemoyer. 2019.
\newblock Bart: Denoising sequence-to-sequence pre-training for natural
  language generation, translation, and comprehension.
\newblock \emph{arXiv preprint arXiv:1910.13461}.

\bibitem[{Li et~al.(2020)Li, Zhou, Xiong, Socher, and Hoi}]{li2020prototypical}
Junnan Li, Pan Zhou, Caiming Xiong, Richard Socher, and Steven~CH Hoi. 2020.
\newblock Prototypical contrastive learning of unsupervised representations.
\newblock \emph{arXiv preprint arXiv:2005.04966}.

\bibitem[{Lloyd(1982)}]{lloyd1982least}
Stuart Lloyd. 1982.
\newblock Least squares quantization in pcm.
\newblock \emph{IEEE transactions on information theory}, 28(2):129--137.

\bibitem[{Ma(2019)}]{ma2019nlpaug}
Edward Ma. 2019.
\newblock Nlp augmentation.
\newblock https://github.com/makcedward/nlpaug.

\bibitem[{Maaten and Hinton(2008)}]{maaten2008visualizing}
Laurens van~der Maaten and Geoffrey Hinton. 2008.
\newblock Visualizing data using t-sne.
\newblock \emph{Journal of machine learning research}, 9(Nov):2579--2605.

\bibitem[{MacQueen et~al.(1967)}]{macqueen1967some}
James MacQueen et~al. 1967.
\newblock Some methods for classification and analysis of multivariate
  observations.
\newblock In \emph{Proceedings of the fifth Berkeley symposium on mathematical
  statistics and probability}, volume~1, pages 281--297. Oakland, CA, USA.

\bibitem[{Meng et~al.(2021)Meng, Xiong, Bajaj, Tiwary, Bennett, Han, and
  Song}]{meng2021coco}
Yu~Meng, Chenyan Xiong, Payal Bajaj, Saurabh Tiwary, Paul Bennett, Jiawei Han,
  and Xia Song. 2021.
\newblock Coco-lm: Correcting and contrasting text sequences for language model
  pretraining.
\newblock \emph{arXiv preprint arXiv:2102.08473}.

\bibitem[{Mikolov et~al.(2013{\natexlab{a}})Mikolov, Chen, Corrado, and
  Dean}]{mikolov2013efficient}
Tomas Mikolov, Kai Chen, Greg Corrado, and Jeffrey Dean. 2013{\natexlab{a}}.
\newblock Efficient estimation of word representations in vector space.
\newblock \emph{arXiv preprint arXiv:1301.3781}.

\bibitem[{Mikolov et~al.(2013{\natexlab{b}})Mikolov, Sutskever, Chen, Corrado,
  and Dean}]{mikolov2013distributed}
Tomas Mikolov, Ilya Sutskever, Kai Chen, Greg~S Corrado, and Jeff Dean.
  2013{\natexlab{b}}.
\newblock Distributed representations of words and phrases and their
  compositionality.
\newblock In \emph{Advances in neural information processing systems}, pages
  3111--3119.

\bibitem[{Morris et~al.(2020)Morris, Lifland, Yoo, Grigsby, Jin, and
  Qi}]{morris2020textattack}
John~X. Morris, Eli Lifland, Jin~Yong Yoo, Jake Grigsby, Di~Jin, and Yanjun Qi.
  2020.
\newblock \href {http://arxiv.org/abs/2005.05909} {Textattack: A framework for
  adversarial attacks, data augmentation, and adversarial training in nlp}.

\bibitem[{Oord et~al.(2018)Oord, Li, and Vinyals}]{oord2018representation}
Aaron van~den Oord, Yazhe Li, and Oriol Vinyals. 2018.
\newblock Representation learning with contrastive predictive coding.
\newblock \emph{arXiv preprint arXiv:1807.03748}.

\bibitem[{Paszke et~al.(2017)Paszke, Gross, Chintala, Chanan, Yang, DeVito,
  Lin, Desmaison, Antiga, and Lerer}]{paszke2017automatic}
Adam Paszke, Sam Gross, Soumith Chintala, Gregory Chanan, Edward Yang, Zachary
  DeVito, Zeming Lin, Alban Desmaison, Luca Antiga, and Adam Lerer. 2017.
\newblock Automatic differentiation in pytorch.
\newblock In \emph{NIPS-W}.

\bibitem[{Peters et~al.(2018)Peters, Neumann, Iyyer, Gardner, Clark, Lee, and
  Zettlemoyer}]{peters2018deep}
Matthew~E Peters, Mark Neumann, Mohit Iyyer, Matt Gardner, Christopher Clark,
  Kenton Lee, and Luke Zettlemoyer. 2018.
\newblock Deep contextualized word representations.
\newblock \emph{arXiv preprint arXiv:1802.05365}.

\bibitem[{Phan et~al.(2008)Phan, Nguyen, and Horiguchi}]{phan2008learning}
Xuan-Hieu Phan, Le-Minh Nguyen, and Susumu Horiguchi. 2008.
\newblock Learning to classify short and sparse text \& web with hidden topics
  from large-scale data collections.
\newblock In \emph{Proceedings of the 17th international conference on World
  Wide Web}, pages 91--100.

\bibitem[{Purushwalkam and Gupta(2020)}]{purushwalkam2020demystifying}
Senthil Purushwalkam and Abhinav Gupta. 2020.
\newblock Demystifying contrastive self-supervised learning: Invariances,
  augmentations and dataset biases.
\newblock \emph{arXiv preprint arXiv:2007.13916}.

\bibitem[{Qu et~al.(2020)Qu, Shen, Shen, Sajeev, Han, and Chen}]{qu2020coda}
Yanru Qu, Dinghan Shen, Yelong Shen, Sandra Sajeev, Jiawei Han, and Weizhu
  Chen. 2020.
\newblock Coda: Contrast-enhanced and diversity-promoting data augmentation for
  natural language understanding.
\newblock \emph{arXiv preprint arXiv:2010.08670}.

\bibitem[{Radford et~al.(2018)Radford, Narasimhan, Salimans, and
  Sutskever}]{radford2018improving}
Alec Radford, Karthik Narasimhan, Tim Salimans, and Ilya Sutskever. 2018.
\newblock Improving language understanding by generative pre-training.

\bibitem[{Rakib et~al.(2020)Rakib, Zeh, Jankowska, and
  Milios}]{rakib2020enhancement}
Md~Rashadul~Hasan Rakib, Norbert Zeh, Magdalena Jankowska, and Evangelos
  Milios. 2020.
\newblock Enhancement of short text clustering by iterative classification.
\newblock \emph{arXiv preprint arXiv:2001.11631}.

\bibitem[{Reimers and
  Gurevych(2019{\natexlab{a}})}]{reimers-2019-sentence-bert}
Nils Reimers and Iryna Gurevych. 2019{\natexlab{a}}.
\newblock \href {http://arxiv.org/abs/1908.10084} {Sentence-bert: Sentence
  embeddings using siamese bert-networks}.
\newblock In \emph{Proceedings of the 2019 Conference on Empirical Methods in
  Natural Language Processing}. Association for Computational Linguistics.

\bibitem[{Reimers and Gurevych(2019{\natexlab{b}})}]{reimers2019sentence}
Nils Reimers and Iryna Gurevych. 2019{\natexlab{b}}.
\newblock Sentence-bert: Sentence embeddings using siamese bert-networks.
\newblock \emph{arXiv preprint arXiv:1908.10084}.

\bibitem[{Ren et~al.(2019)Ren, Deng, He, and Che}]{ren2019generating}
Shuhuai Ren, Yihe Deng, Kun He, and Wanxiang Che. 2019.
\newblock Generating natural language adversarial examples through probability
  weighted word saliency.
\newblock In \emph{Proceedings of the 57th annual meeting of the association
  for computational linguistics}, pages 1085--1097.

\bibitem[{Sebrechts et~al.(1999)Sebrechts, Cugini, Laskowski, Vasilakis, and
  Miller}]{sebrechts1999visualization}
Marc~M Sebrechts, John~V Cugini, Sharon~J Laskowski, Joanna Vasilakis, and
  Michael~S Miller. 1999.
\newblock Visualization of search results: a comparative evaluation of text,
  2d, and 3d interfaces.
\newblock In \emph{Proceedings of the 22nd annual international ACM SIGIR
  conference on Research and development in information retrieval}, pages
  3--10.

\bibitem[{Shaham et~al.(2018)Shaham, Stanton, Li, Nadler, Basri, and
  Kluger}]{shaham2018spectralnet}
Uri Shaham, Kelly Stanton, Henry Li, Boaz Nadler, Ronen Basri, and Yuval
  Kluger. 2018.
\newblock Spectralnet: Spectral clustering using deep neural networks.
\newblock \emph{arXiv preprint arXiv:1801.01587}.

\bibitem[{Shen et~al.(2019)Shen, Ott, Auli, and Ranzato}]{shen2019mixture}
Tianxiao Shen, Myle Ott, Michael Auli, and Marc'Aurelio Ranzato. 2019.
\newblock Mixture models for diverse machine translation: Tricks of the trade.
\newblock \emph{International Conference on Machine Learning}.

\bibitem[{Van Der~Maaten(2009)}]{van2009learning}
Laurens Van Der~Maaten. 2009.
\newblock Learning a parametric embedding by preserving local structure.
\newblock In \emph{Artificial Intelligence and Statistics}, pages 384--391.

\bibitem[{Wu et~al.(2018)Wu, Xiong, Yu, and Lin}]{wu2018unsupervised}
Zhirong Wu, Yuanjun Xiong, Stella~X Yu, and Dahua Lin. 2018.
\newblock Unsupervised feature learning via non-parametric instance
  discrimination.
\newblock In \emph{Proceedings of the IEEE Conference on Computer Vision and
  Pattern Recognition}, pages 3733--3742.

\bibitem[{Xie et~al.(2016)Xie, Girshick, and Farhadi}]{xie2016unsupervised}
Junyuan Xie, Ross Girshick, and Ali Farhadi. 2016.
\newblock Unsupervised deep embedding for clustering analysis.
\newblock In \emph{International conference on machine learning}, pages
  478--487.

\bibitem[{Xu et~al.(2017)Xu, Xu, Wang, Zheng, Tian, and Zhao}]{xu2017self}
Jiaming Xu, Bo~Xu, Peng Wang, Suncong Zheng, Guanhua Tian, and Jun Zhao. 2017.
\newblock Self-taught convolutional neural networks for short text clustering.
\newblock \emph{Neural Networks}, 88:22--31.

\bibitem[{Yin and Wang(2016)}]{yin2016model}
Jianhua Yin and Jianyong Wang. 2016.
\newblock A model-based approach for text clustering with outlier detection.
\newblock In \emph{2016 IEEE 32nd International Conference on Data Engineering
  (ICDE)}, pages 625--636. IEEE.

\bibitem[{Zhang et~al.(2017{\natexlab{a}})Zhang, Sun, Eriksson, and
  Balzano}]{zhang2017deep}
Dejiao Zhang, Yifan Sun, Brian Eriksson, and Laura Balzano. 2017{\natexlab{a}}.
\newblock Deep unsupervised clustering using mixture of autoencoders.
\newblock \emph{arXiv preprint arXiv:1712.07788}.

\bibitem[{Zhang et~al.(2017{\natexlab{b}})Zhang, Cisse, Dauphin, and
  Lopez-Paz}]{zhang2017mixup}
Hongyi Zhang, Moustapha Cisse, Yann~N Dauphin, and David Lopez-Paz.
  2017{\natexlab{b}}.
\newblock mixup: Beyond empirical risk minimization.
\newblock \emph{arXiv preprint arXiv:1710.09412}.

\bibitem[{Zhang et~al.(2016)Zhang, Isola, and Efros}]{zhang2016colorful}
Richard Zhang, Phillip Isola, and Alexei~A Efros. 2016.
\newblock Colorful image colorization.
\newblock In \emph{European conference on computer vision}, pages 649--666.
  Springer.

\bibitem[{Zhang and LeCun(2015)}]{zhang2015text}
Xiang Zhang and Yann LeCun. 2015.
\newblock Text understanding from scratch.
\newblock \emph{arXiv preprint arXiv:1502.01710}.

\end{thebibliography}
\bibliographystyle{acl_natbib}

\appendix
\section{Appendices}
\label{sec:appendix}

\begin{figure*}[!htbp]
    \centering
    \includegraphics[scale=0.38]{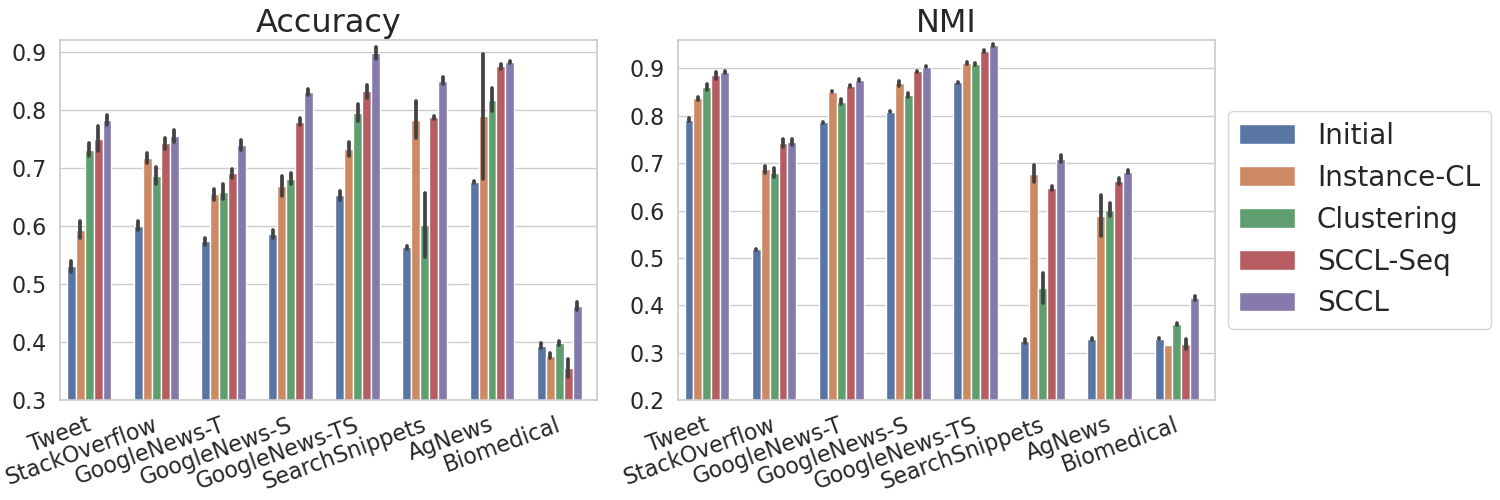}
    \caption{Ablation study of our proposed SCCL model. In SCCL-Seq, we first train the model using Instance-CL, and then optimize the clustering objective. }
    \label{fig:ablation_study_supplement}
\end{figure*}

\subsection{Experiment Setup}
\label{appendix:exp_setup}
We use the Adam optimizer \cite{KingmaB14} with batch size of 400. We use  \textit{distilbert-base-nli-stsb-mean-tokens} in the Sentence Transformers library \citep{reimers-2019-sentence-bert} as the backbone, and we set the maximum input length to 32. We use a constant learning rate 5e-6 to optimize the backbone, while setting learning rate to 5e-4 to optimizing both the Clustering head and Instance-CL head. Same as \citet{xie2016unsupervised, hadifar2019self}, we set $\alpha=1$ for all datasets except Biomedical where we use $\alpha=10$. As mentioned in Section \ref{sec:model}, we set $\tau=0.5$ for optimizing the constrastive loss. We tried different $\tau$ values in the range of $(0,1]$ and found using $\tau=0.5$ yields comparatively better yet stable performance across datasets for both Instance-CL and SCCL. For fair comparison between SCCL and its components or variants, we report the clustering performance for each of them by applying KMeans on the representations post the associated training processes.

\subsection{Data Augmentation}
\label{appendix:var_aug}
\paragraph{Exploration of Data Augmentations}
As mentioned in Section \ref{subsec:var_aug}, we tried three different augmentation strengths for both \textit{WordNet Augmenter} and \textit{Contextual Augmenter} by choosing the word substitution ratio of each text instance as $10\%$, $20\%$, and $30\%$, respectively. For each augmentation strength, we generate a pair of augmentations for each text instance. As for \textit{Paraphrase via back translation}, we computed the BLEU score between each original instance and its ten candidate paraphrases. We then select three pairs, achieving the highest, medium, and lowest BLEU scores, from the ten condidates as the augmented data. For each augmentation  method, we run SCCL on all three augmentation strengths independently and report the best result. 

For both \textit{WordNet Augmenter} and \textit{Contextual Augmenter}, we observe that comparatively longer text instances, \ie those in AgNews, SearchSnippets, GoogleNewsTS, and GoogleNewsS, benefit from stronger augmentation. 
In contrast, \textit{Paraphrase via back translation} shows better results when evaluated on the augmented pairs achieving the lowest BLEU scores with the original instance, \ie the pair achieving the two lowest scores among all ten condidate paraphrases for each text instance. 

\paragraph{Building Effective Data Augmentations for NLP} As discussed in Section \ref{subsec:comp_dataaug}, using composition of data augmentations is not always beneficial for short text clustering. Because changing a single word can invert the meaning of the whole sentence, and the challenge is compounded when applied a second round data augmentation to short text data. However, we would hopefully cross the hurdle soon, as more effective approaches are keeping developed by the NLP community \citet{qu2020coda, giorgi2020declutr, meng2021coco}.

\subsection{Alternative Clustering Loss for SCCL}
\label{appendix_cluster}
In the current form of SCCL, the clustering loss is optimized on the original dataset only. However, several alternatives could be considered, we discuss two options here to encourage further explorations. 

\paragraph{Alternative 1.}{Let $j^1$ and $j^2$ denote the indices of the augmented pair for the $j^{th}$ text instance in the original set, respectively. For the augmented instance $j^1$, we then push the cluster assignment probability towards the target distribution obtained by the other instance $j^2$, and vice versa.  That is, we replace Eq (\ref{eq:KL_cluster_instance}) with the following  
\begin{align}
    \ell_j^C = \text{KL}\left[p_{j^1} || q_{j^2}\right] + \text{KL}\left[p_{j^2} || q_{j^1}\right]
    \label{eq:alternative_1}
\end{align}
Here $p$ and $q$ denote the target distribution and the cluster assignment probability defined in Eqs (\ref{eq:cluster_assignment}) and (\ref{eq:target}), respectively.
}

\paragraph{Alternative 2.}{Let $j^0$ and $j^1, j^2$ denote the indices of the original text instance and its augmented pair, respectively. We then use the original instance as anchor, and push the cluster assignments of the augmented pair towards it by optimizing the following
\begin{align}
    \ell_j^C = \text{KL}\left[p_{j^0} || q_{j^1}\right] + \text{KL}\left[p_{j^0} || q_{j^2}\right]
    \label{eq:alternative_2}
\end{align}

Exploring (\ref{eq:alternative_1}) and (\ref{eq:alternative_2}) is out of the scope of this paper, however, it's worth trying when applying SCCL to solve different application problems. Especially considering that the above alternatives might lead to further performance improvement by jointly optimizing the instance-level and the cluster assignment level contrastive learning losses. 
}

\subsection{Supplement materials for ablation study}
\label{appdix:ablation}
Figure \ref{fig:ablation_study_supplement} provides the full version of Figure \ref{fig:ablation_study} in Section \ref{sec:numerical}.

\subsection{Comparison with \citet{rakib2020enhancement}}
\label{appendix_rakib}
While \citet{rakib2020enhancement} achieve better Accuracy on Tweet and GoogleNews-T, we highlight the scalability of our model to large scale data, whereas \citet{rakib2020enhancement} depend on agglomerative clustering which often suffers from high computation complexity. Specifically, let $N$ denote the number of training examples, and $K$ denote the number of clusters. The HAC-SD method proposed by \citet{rakib2020enhancement} first computes the pairwise similarity among all possible pairs of the data, and then sorts the $N^2$ similarity values so as to select the top $N^2/K$ pairwise similarity as the input to the agglomerative clustering algorithm. Thereby, before clustering, HAC-SD could result in $O(N^2\log N)$ time complexity, and  $O(N^2/K)$ storage complexity. Moreover, the agglomerative clustering algorithm could require $O(N^2\log(N/K))$ time complexity. Therefore, HAC-SD is less feasible in presence of large scale data. In contrast, SCCL performs standard stochastic optimization, the time complexity linearly scales with $N$ since SCCL often requires $20-100$ epochs to converge, which is often much smaller than the number of data examples. 


\end{document}